\documentclass{article}
\usepackage{spconf,amsmath,epsfig}
\usepackage{tikz}
\usepackage{multirow}
\usepackage{makecell}
\usetikzlibrary{shapes.geometric, arrows}

\tikzstyle{process} = [rectangle, rounded corners, minimum width=3cm, minimum height=1cm, text centered, draw=black, fill=blue!20]
\tikzstyle{arrow} = [thick,->,>=stealth]

\usepackage{subcaption} 

\providecommand{\doi}[1]{doi: {\footnotesize \href{http://dx.doi.org/#1}{\path{#1}}}}

\usepackage[pdftex=true,breaklinks=true,hidelinks=true,colorlinks=true,citecolor=blue]{hyperref}

\usepackage[square,numbers]{natbib}

\title{Real-Time Blind Defocus Deblurring for Earth Observation: The IMAGIN-e Mission Approach}
\name{Alejandro D. Mousist}
\address{Thales Alenia Space, Tres Cantos, Spain}
\begin{document}
\ninept
\maketitle
%
\begin{abstract}
This work addresses mechanical defocus in Earth observation images from the IMAGIN-e mission aboard the ISS, proposing a blind deblurring approach adapted to space-based edge computing constraints. Leveraging Sentinel-2 data, our method estimates the defocus kernel and trains a restoration model within a GAN framework, effectively operating without reference images.

On Sentinel-2 images with synthetic degradation, SSIM improved by 72.47\% and PSNR by 25.00\%, confirming the model's ability to recover lost details when the original clean image is known. On IMAGIN-e, where no reference images exist, \textbf{perceptual quality metrics indicate a substantial enhancement}, with \textbf{NIQE improving by 60.66\%} and \textbf{BRISQUE by 48.38\%},  \textbf{validating real-world onboard restoration}.  The approach is currently deployed aboard the IMAGIN-e mission, demonstrating its practical application in an operational space environment.

By efficiently handling high-resolution images under edge computing constraints, the method enables applications such as water body segmentation and contour detection while maintaining processing viability despite resource limitations.

\end{abstract}

%

\section{INTRODUCTION AND STATE-OF-THE-ART}
\label{sec:intro}
\begin{figure}
  \centering
  \includegraphics[width=0.66\columnwidth]{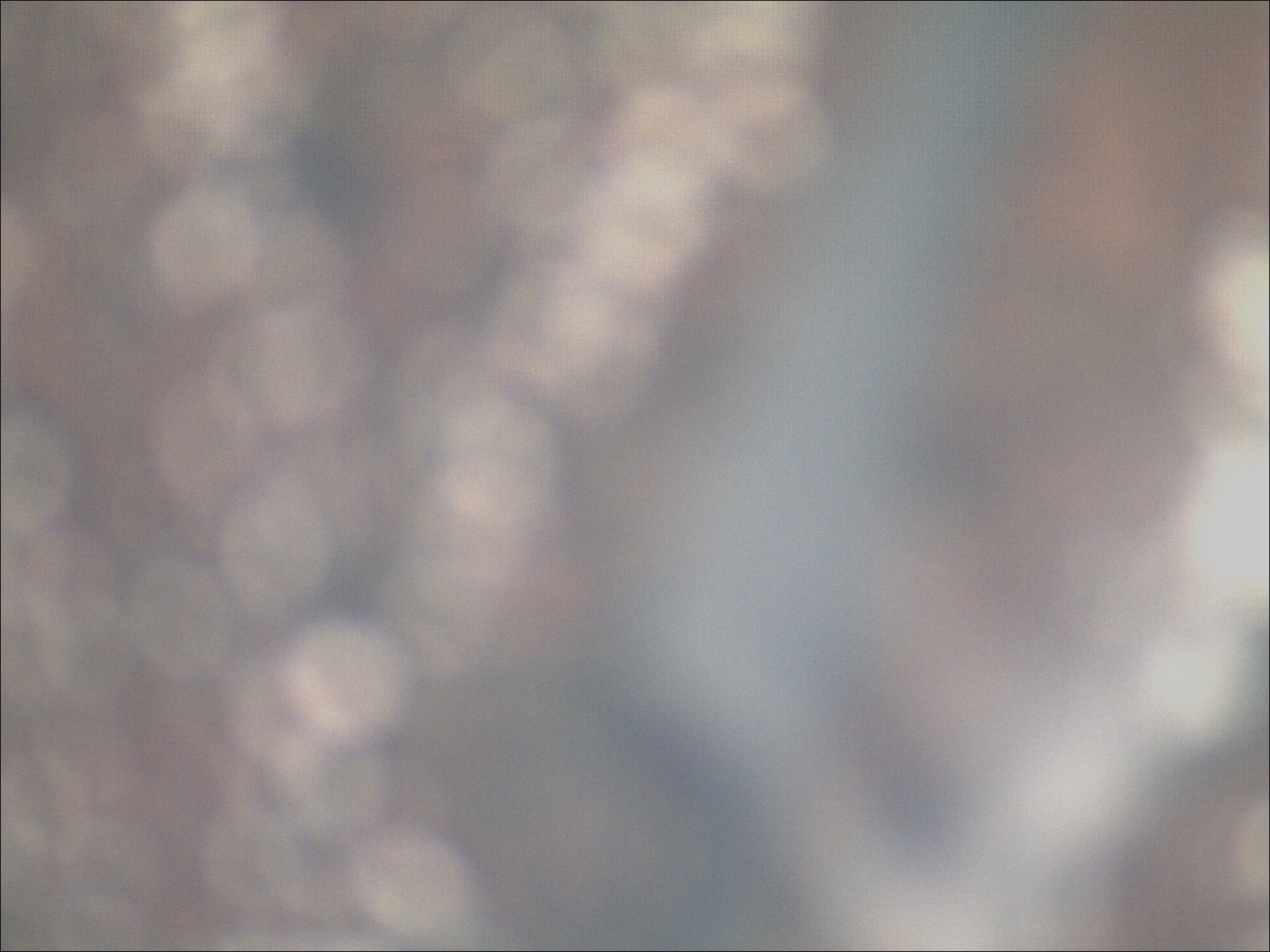}
  \caption{\label{blurred_11} Captured image from the IMAGIN-e payload without further processing, showing significant mechanical blur.}
\end{figure}

The IMAGIN-e mission (\textbf{I}SS \textbf{M}ounted \textbf{A}ccessible \textbf{G}lobal \textbf{I}maging \textbf{N}od\textbf{-e}) is a space edge computing initiative hosted aboard the International Space Station (ISS). IMAGIN-e operates as a functional demonstration payload with real-world applications for Earth observation. Its primary objective is to evaluate the capabilities and operating modes of onboard edge computing by processing Earth observation data directly within the payload. An optical sensor was integrated to capture images that fuel onboard applications. However, the captured images exhibit significant mechanical defocus characterized by wide dispersion and smoothing (see Fig.~\ref{blurred_11}), complicating precise interpretation and hindering the extraction of meaningful insights.

In this context, missions like Sentinel-2 from the Copernicus program -which provide multispectral images with higher spatial resolution (GSD) and additional spectral bands— could serve as a reference to estimate the defocus kernel when contrasted with IMAGIN-e RGB images. Nonetheless, IMAGIN-e images are not georeferenced at origin and include uncertainties (e.g., the sensor's final orientation due to its uncharacterized mechanical and thermoelastic misalignments), posing a significant challenge for restoration in the absence of sharp reference images.

Recent studies, such as Popika and Lelechenko \cite{popika2024machine}, have used synthetic distortions to train models for satellite image restoration in post-processing. Our approach builds on this idea, adapting it for onboard edge computing to enable real-time correction within the IMAGIN-e payload (see Section~\ref{sec:methodology}).

Traditional deblurring approaches, such as the Wiener filter \cite{wiener1949extrapolation} or Richardson-Lucy deconvolution \cite{richardson1972bayesian}, rely on known blur kernel characteristics, which limits their performance for complex, non-uniform blurs observed in space-based imagery. 
The advent of deep learning has enabled robust alternative strategies. Early methods employed Convolutional Neural Networks (CNNs) to learn the mapping between blurred and sharp images \cite{nah2017deep, tao2018scale}, while GAN‑based approaches like DeblurGAN \cite{kupyn2018deblurgan,kupyn2019deblurgan} addressed blind deblurring when the blur kernel is unknown. More recently, transformer‑based architectures have emerged as promising candidates for image restoration tasks. For instance, DeblurDiNAT\cite{liu2024deblurdinat} presents a compact model that leverages dilated neighborhood attention mechanisms to achieve robust generalization and high perceptual fidelity, even in out‑of‑domain settings . In parallel, MIMO‑Uformer \cite{zhang2024mimo} integrates a U‑shaped structure with window‐based attention (W‑MSA), enabling efficient capture of both local and global dependencies with a computational footprint suitable for resource‐constrained environments.

Despite these advances, most state‑of‑the‑art approaches assume access to paired blurred‑sharp images or mandate substantial computational resources, rendering them incompatible with the onboard processing constraints of the IMAGIN‑e mission. 

\subsection{Contribution of This Work}
Our research contributes a blind deblurring methodology for satellite imagery without reference images that leverages Sentinel-2 data to characterize the defocus kernel. We adapt MIMO-Unet++\cite{mimounet} for space-based edge computing, optimizing computational efficiency while preserving restoration quality. Quantitative and qualitative analysis validates our approach, showing significant improvements in structural similarity and edge preservation. Additionally, we provide insights into deep learning-based image enhancement for space-based observation systems with limited resources.

This study introduces a generative AI framework for defocus correction within the constraints of the IMAGIN-e mission, enhancing onboard edge computing for Earth observation and enabling the effective utilization of otherwise compromised instruments.

\section{Problem Characterization}
\label{sec:methods}

\subsection{Platform and Payload Orientation}

\begin{figure}
  \centering
  \includegraphics[width=0.7\columnwidth]{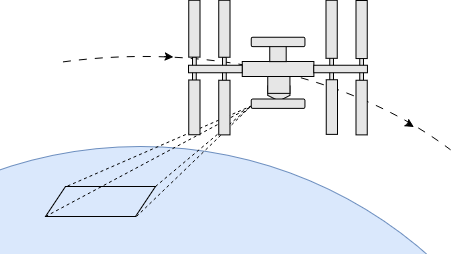}
  \caption{\label{iss} Illustration of the payload orientation on the Bartolomeo platform, showing its backward tilt relative to the ISS trajectory.}
\end{figure}

The payload is hosted on an external platform for payload hosting, mounted on the Columbus module, which is externally mounted on the Columbus module of the International Space Station (ISS). Although its nominal alignment is Earth-facing, the imaging system is not perfectly oriented in the nadir direction; rather, it is directed a few degrees backward relative to the ISS trajectory (see Fig. \ref{iss}). This orientation results in a non-perpendicular incidence angle compared to a purely nadir-pointing configuration, potentially affecting the observation geometry and data acquisition characteristics. Moreover, the payload was installed  using a robotic arm, so the exact sensor orientation relative to nadir was not known a priori.

\subsection{Sensor Data Characteristics}

\begin{figure}
    \centering
    \begin{subfigure}{0.9\columnwidth}
        \centering
        \includegraphics[width=\textwidth]{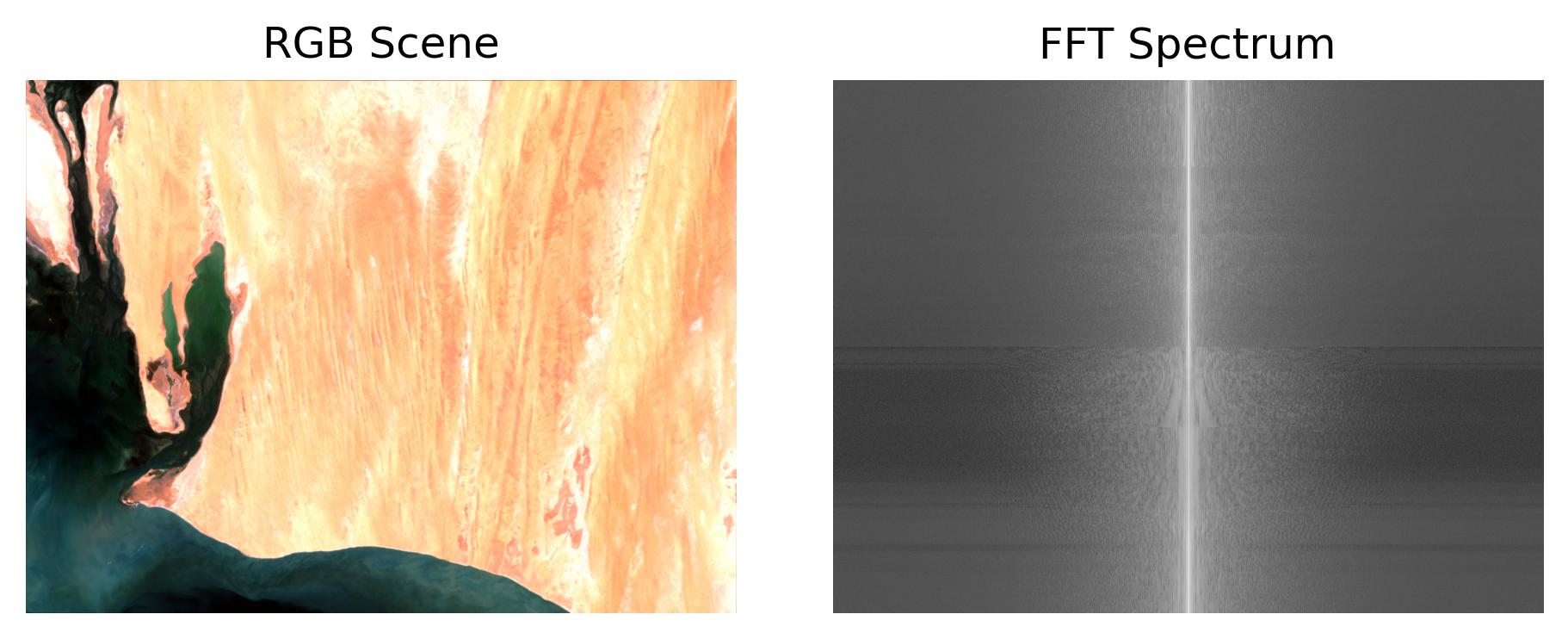}
        \caption{Sentinel-2: Scene and FFT spectrum}
        \label{fig:sentinel2}
    \end{subfigure}
    
    \begin{subfigure}{0.9\columnwidth}
        \centering
        \includegraphics[width=\textwidth]{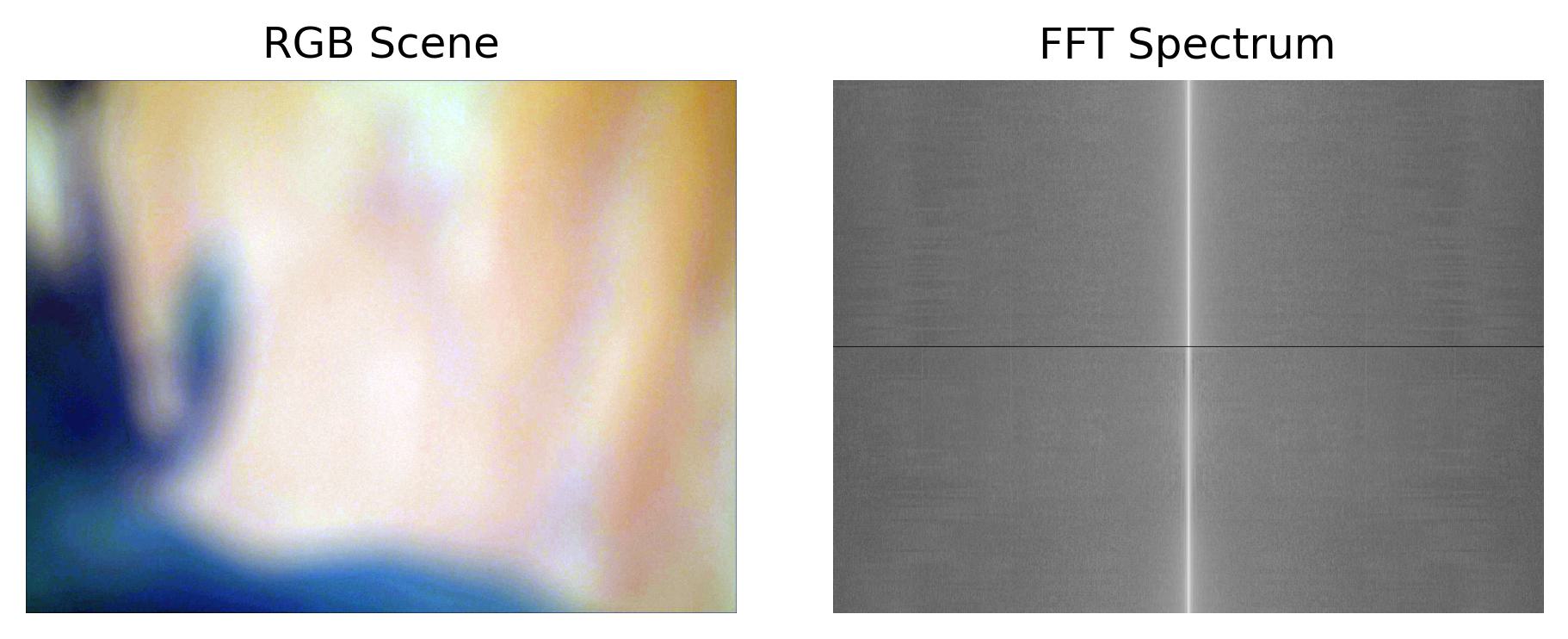}
        \caption{IMAGIN-e: Scene and FFT spectrum}
        \label{fig:imagine}
    \end{subfigure}

    \caption{\label{fft}Comparison of Sentinel-2 and IMAGIN-e images along with their frequency spectra. The Sentinel-2 scene, composed of RGB bands downscaled to a 40m GSD, and its corresponding frequency spectrum are presented in (a). The IMAGIN-e scene and its respective frequency spectrum are shown in (b), illustrating the effects of defocus and alterations in the frequency domain.}
    \label{fig:comparacion}
\end{figure}

The sensor acquires RGB images compressed in JPEG format at a resolution of 2048×1536 pixels. The Ground Sample Distance (GSD) ranges from 37.5m to 41m, depending on altitude variations, ISS pitch fluctuations, and terrain elevation changes. The captured images exhibit significant optical defocus noise, likely due to mechanical miscalibration, while some images also display minor shot noise, though its intensity is considerably lower than that of the defocus blur. Figure \ref{fft} provides a spectral comparison between an IMAGIN-e capture and its corresponding Sentinel-2 scene, highlighting the frequency-domain effects of these noise sources.

\subsection{Onboard Deblurring Process}

The deblurring process is designed to be executed onboard without dedicated acceleration hardware as a critical step in the postprocessing stage of the capture pipeline. It takes place immediately after image acquisition, ensuring that restoration is completed before the images are passed on for further analysis. Third-party applications, which request image captures and process them upon availability, rely on this preprocessing step to enhance data quality and optimize downstream computational tasks.

Given the constraints of onboard execution without specialized hardware, the deblurring model must operate efficiently within the platform’s limited computational resources. To meet this challenge, the MIMO-Unet++ model was selected for its high efficiency in generative processing, enabling real-time deblurring with minimal hardware requirements. By integrating this model into the capture pipeline, image restoration is performed onboard without compromising system performance, ensuring that the processed images maintain the necessary fidelity for further analysis.

\section{Methodology: Deblurring without reference images}
\label{sec:methodology}

\subsection{Model Architecture and Training Strategy}
To enhance structural features critical for georeferencing, we extracted 1024×1024 pixel patches from Sentinel-2 imagery and downscaled them to 256×256 pixels. This size reduction simplified the learning process by focusing the model on sharpening primary edge structures rather than on subtle textures. A batch size of 4 patches was chosen to balance computational efficiency with training stability. We used a MultiStepLR schedule with an initial learning rate of 1e-4, reducing it every 500 iterations by a factor of 0.5 over 3000 iterations to progressively refine the model’s ability to produce spatially coherent reconstructions.

Initially, only defocused images—accompanied by tentative geolocation from the ISS’s position and attitude data were available, making it extremely difficult to align these images with established ground references due to severe defocus and unknown noise characteristics. To tackle this, we first trained an early version of the MIMO-Unet++ model using RGB images generated from Sentinel2 products and augmented with various noise types (Gaussian, defocus, shot, motion, and spin blur). The outputs of this model allowed us to correlate the images relative to their Sentinel-2 counterparts, leading to improved noise characterization and the creation of more realistic synthetic training data.

Subsequently, we used these synthetic images to train a refined MIMO-Unet++ model within a GAN framework, with the model serving as the generator. A multi-scale discriminator was employed to leverage the generator’s outputs at different scales inspired by Pix2pixHD\cite{wang2018highresolution}, enhanced with self-attention mechanisms \cite{zhang2019selfattention} and spectral normalization, ensuring effective extraction of features across all resolutions and promoting superior image reconstruction. 
The overall loss function combined the standard adversarial loss with an L1 loss and an FFT-domain loss—as proposed in the original MIMO-Unet++ framework—as well as a perceptual loss computed using a VGG16 model pre-trained on Sentinel-2 images. This comprehensive training strategy yielded a robust generator capable of delivering deblurred images with enhanced visual fidelity and structural accuracy, which is crucial for Earth observation tasks in edge computing environments.

\subsection{Edge Implementation}
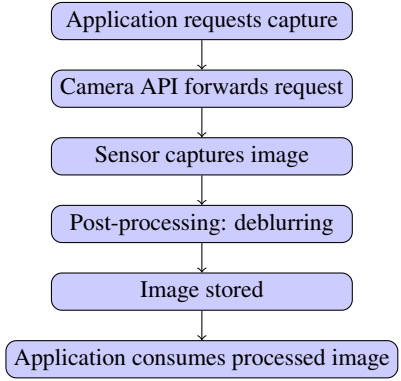
\begin{figure}[t]
    \centering
    \begin{tikzpicture}[
        node distance=0.9cm,
        every node/.style={draw, rounded corners, text centered, minimum width=4cm, minimum height=0.4cm, fill=blue!20},
        every arrow/.style={thick,->,>=stealth}
    ]

        \node (app) {Application requests capture};
        \node (api) [below of=app] {Camera API forwards request};
        \node (sensor) [below of=api] {Sensor captures image};
        \node (postprocess) [below of=sensor] {Post-processing: deblurring};
        \node (storage) [below of=postprocess] {Image stored};
        \node (consume) [below of=storage] {Application consumes processed image};

        \draw [->] (app) -- (api);
        \draw [->] (api) -- (sensor);
        \draw [->] (sensor) -- (postprocess);
        \draw [->] (postprocess) -- (storage);
        \draw [->] (storage) -- (consume);

    \end{tikzpicture}
    \caption{The diagram illustrates the position of the deblurring process within the image processing chain. An application requests an image from the camera API, which then communicates with the sensor for acquisition. The raw image undergoes a post-processing stage, including deblurring, before being stored for later consumption by the application.}
    \label{fig:processing_chain}
\end{figure}

For deployment in the IMAGIN-e mission, the model must operate onboard a hosted payload on the ISS, sharing computational resources with other processes and without dedicated acceleration hardware. Therefore, it is imperative to maintain low latency to ensure seamless integration into the image post-processing pipeline (see Fig.~\ref{fig:processing_chain}). The system constraints summarized in Table~\ref{tab:conditions} require that processing speed and resource usage be carefully managed to meet the rigorous demands of edge computing environments.

\begin{table}
    \caption{Problem conditions}
    \label{tab:conditions}
    \centering
    \begin{tabular}{|l|c|}
        \hline \hline
        \textbf{Parameter} & \textbf{Value} \\
        \hline
        Acceleration HW & Not present \\
        Available RAM memory & 300 MB \\
        Virtual memory & 2 GB \\
        Available CPU & 3 cores (shared) \\
        \hline \hline
    \end{tabular}
\end{table}


\section{Results and Discussion}
\label{sec:results}

\begin{figure}
    \centering
    \begin{subfigure}[t]{0.47\columnwidth}
        \centering
        \includegraphics[width=\linewidth]{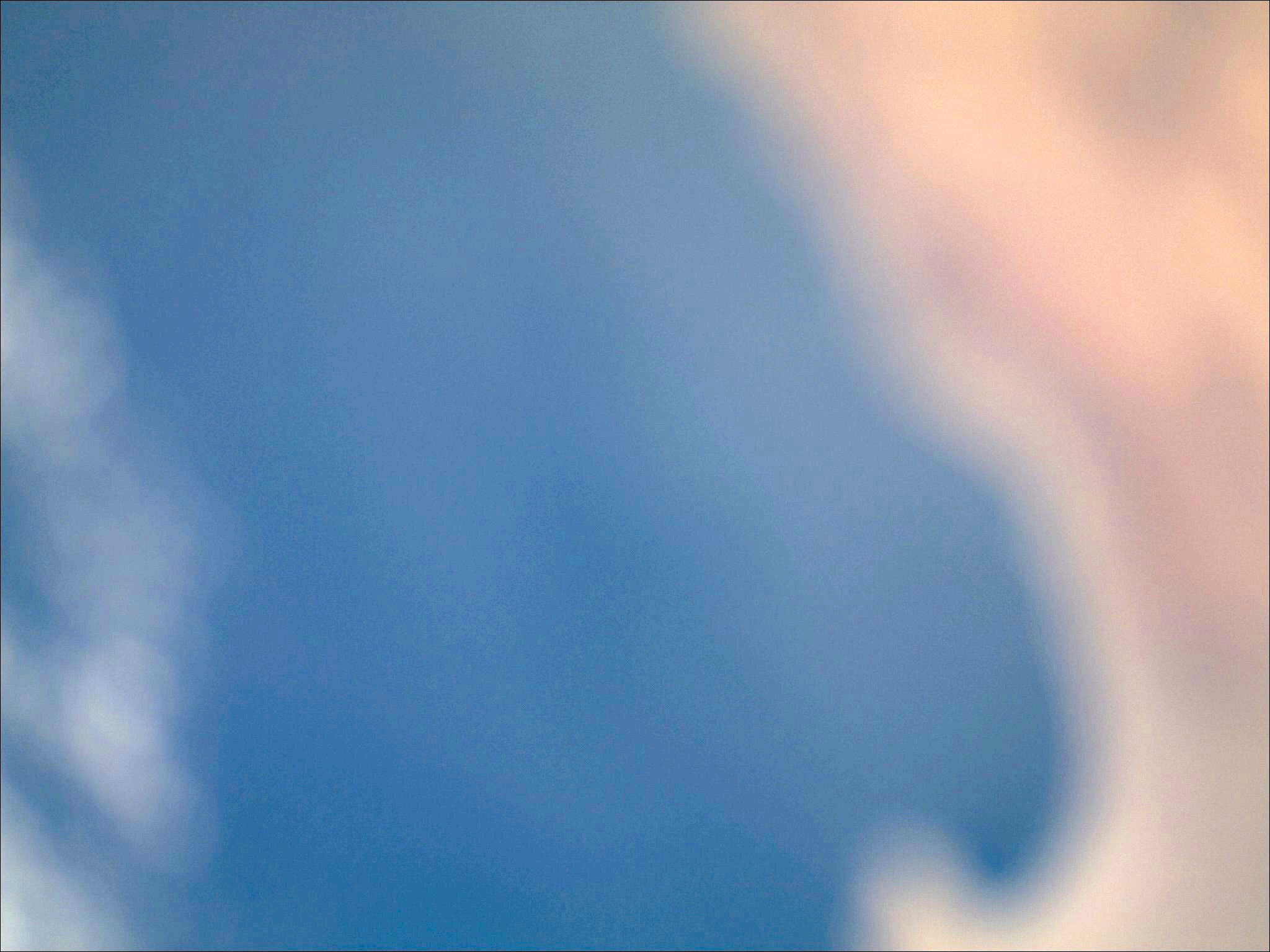}
        \caption{Raw image}
        \label{fig:raw}
    \end{subfigure}
    \hfill
    \begin{subfigure}[t]{0.47\columnwidth}
        \centering
        \includegraphics[width=\linewidth]{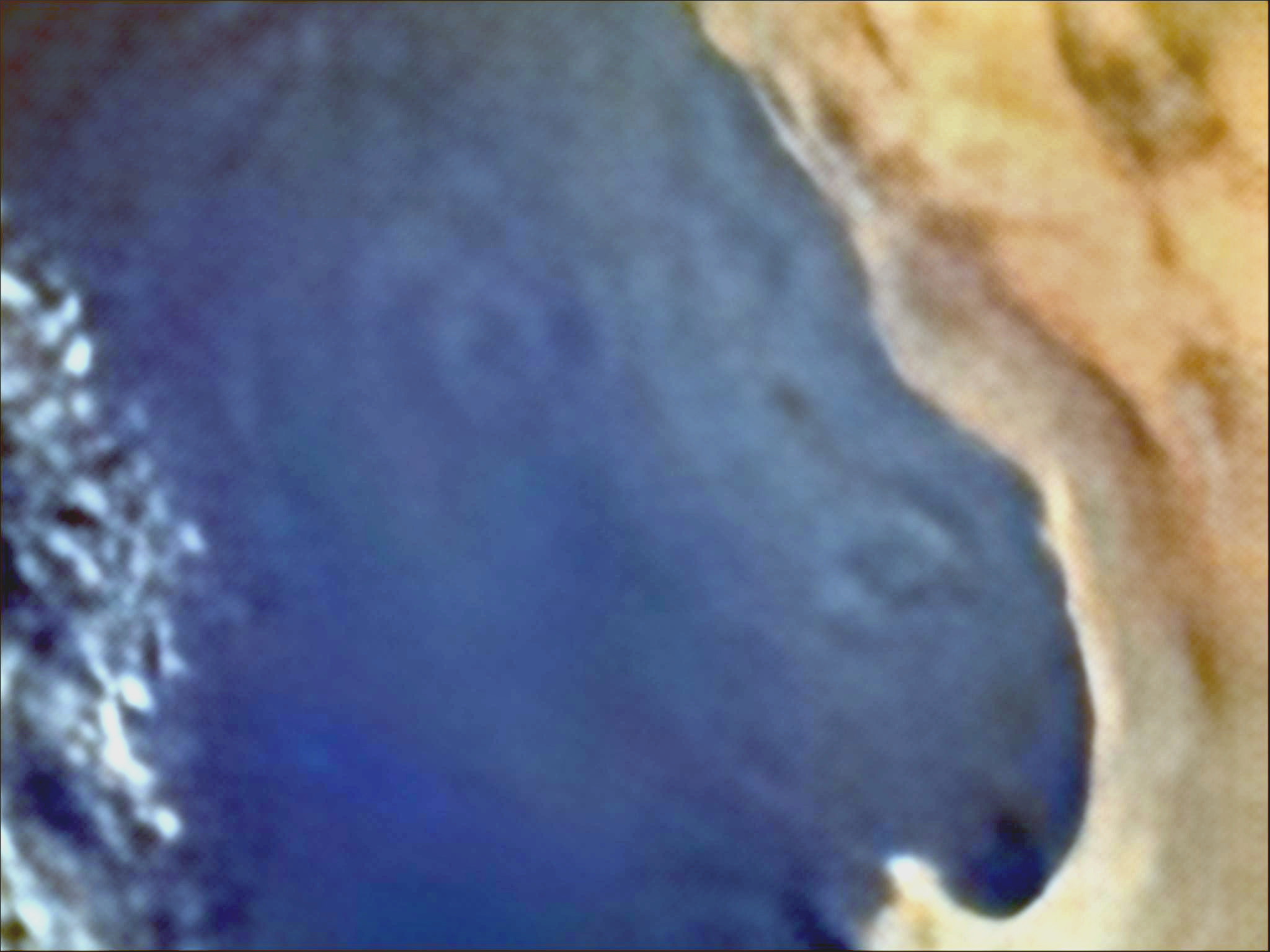}
        \caption{Deblurred image}
        \label{fig:processed}
    \end{subfigure}
    \caption{Initial deblurring effectively sharpened main borders but produced low quality images and ringing effect on some captures. Left image (Fig.\ref{fig:raw}) shows the output of the sensor, while right image (Fig.\ref{fig:processed}) shows the deblurred scene with the initial model.}
    \label{fig:initial_model}
\end{figure}

\begin{figure}
    \centering
    \begin{subfigure}{\columnwidth}
        \begin{subfigure}[t]{0.47\columnwidth}
            \centering
            \includegraphics[width=\linewidth]{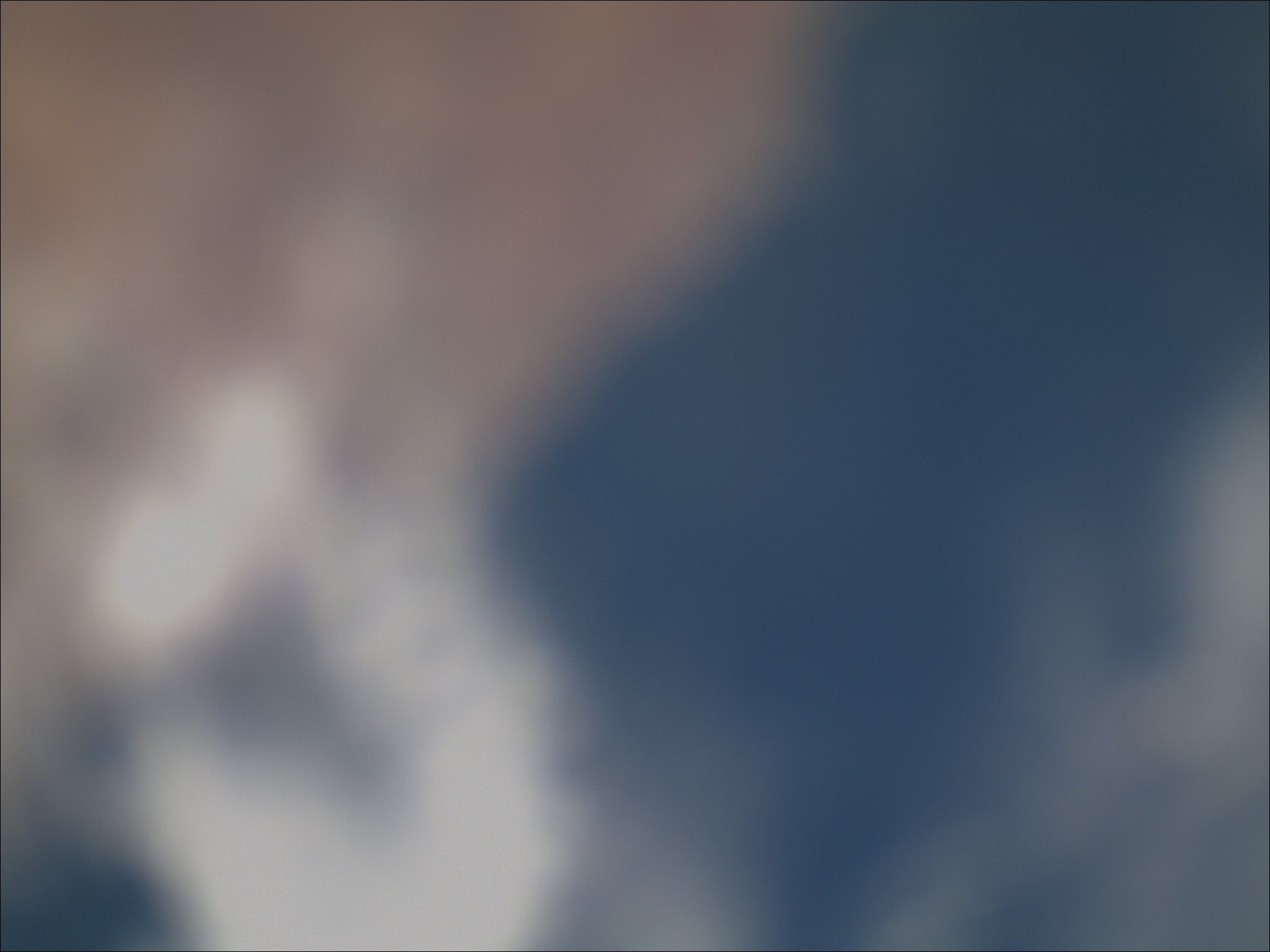}
        \end{subfigure}
        \hfill
        \begin{subfigure}[t]{0.47\columnwidth}
            \centering
            \includegraphics[width=\linewidth]{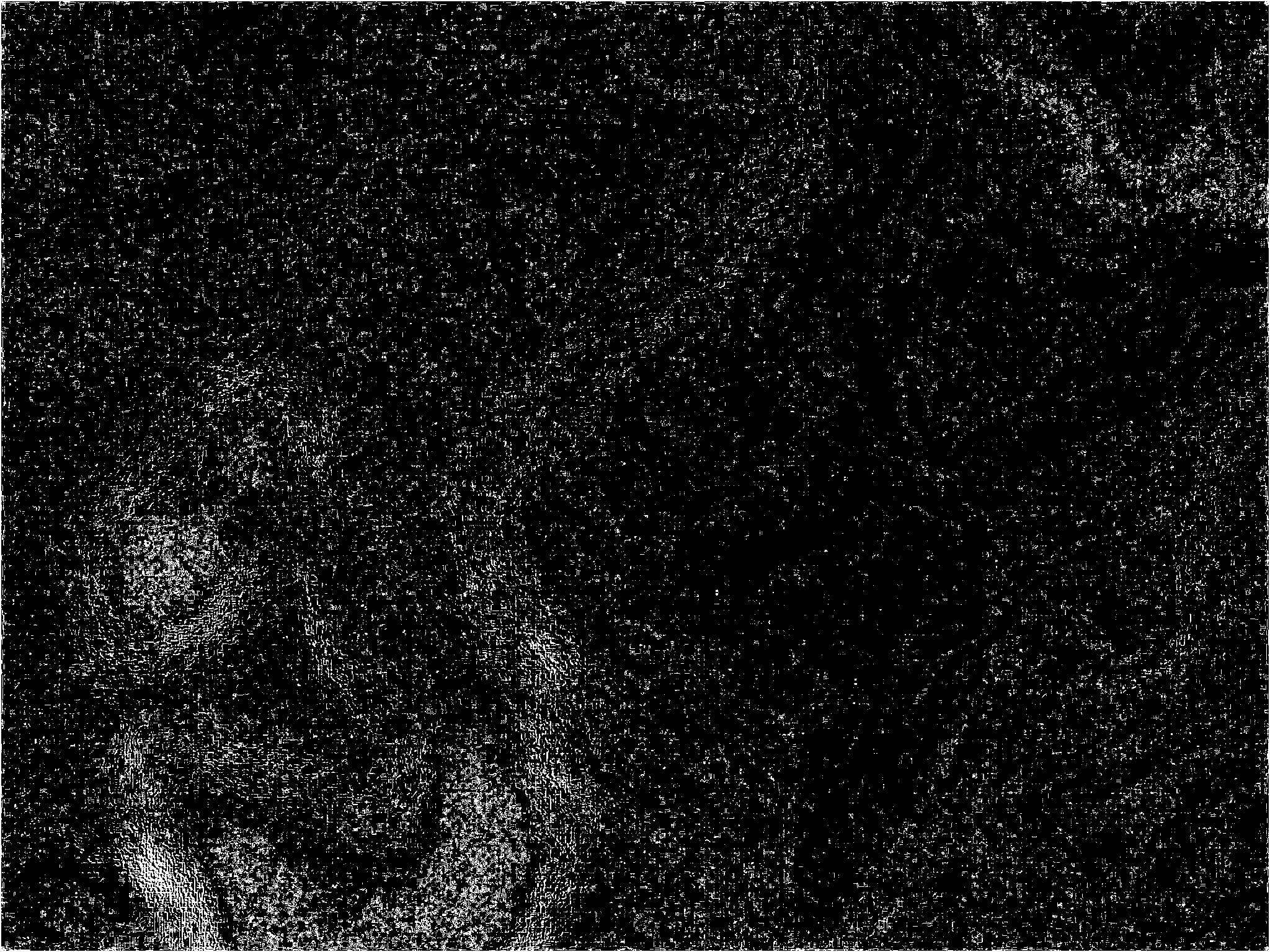}
        \end{subfigure}
        \caption{Border extraction from raw image}
        \label{fig:borders_raw}
    \end{subfigure}
    \begin{subfigure}{\columnwidth}
        \begin{subfigure}[t]{0.47\columnwidth}
            \centering
            \includegraphics[width=\linewidth]{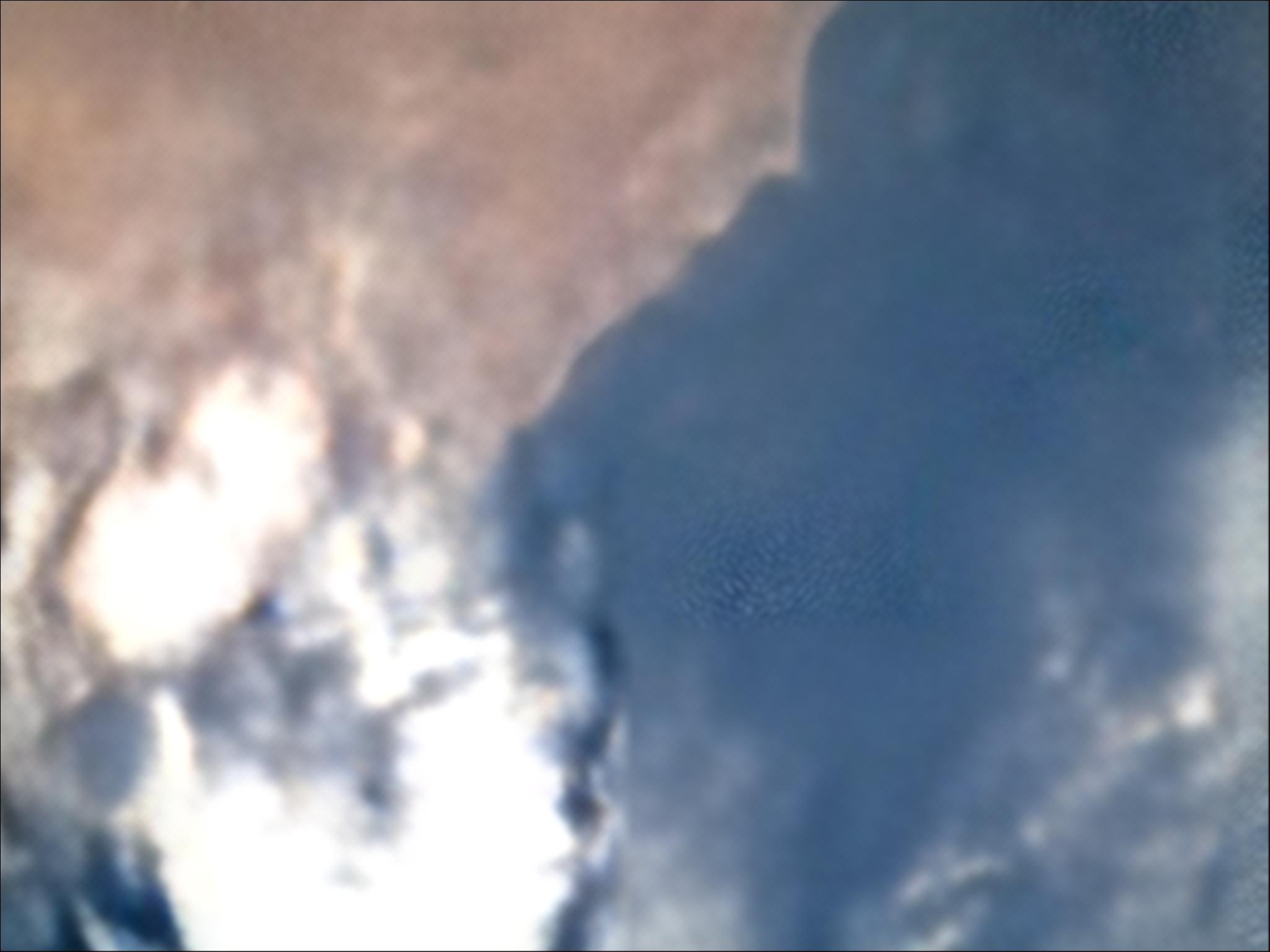}
        \end{subfigure}
        \hfill
        \begin{subfigure}[t]{0.47\columnwidth}
            \centering
            \includegraphics[width=\linewidth]{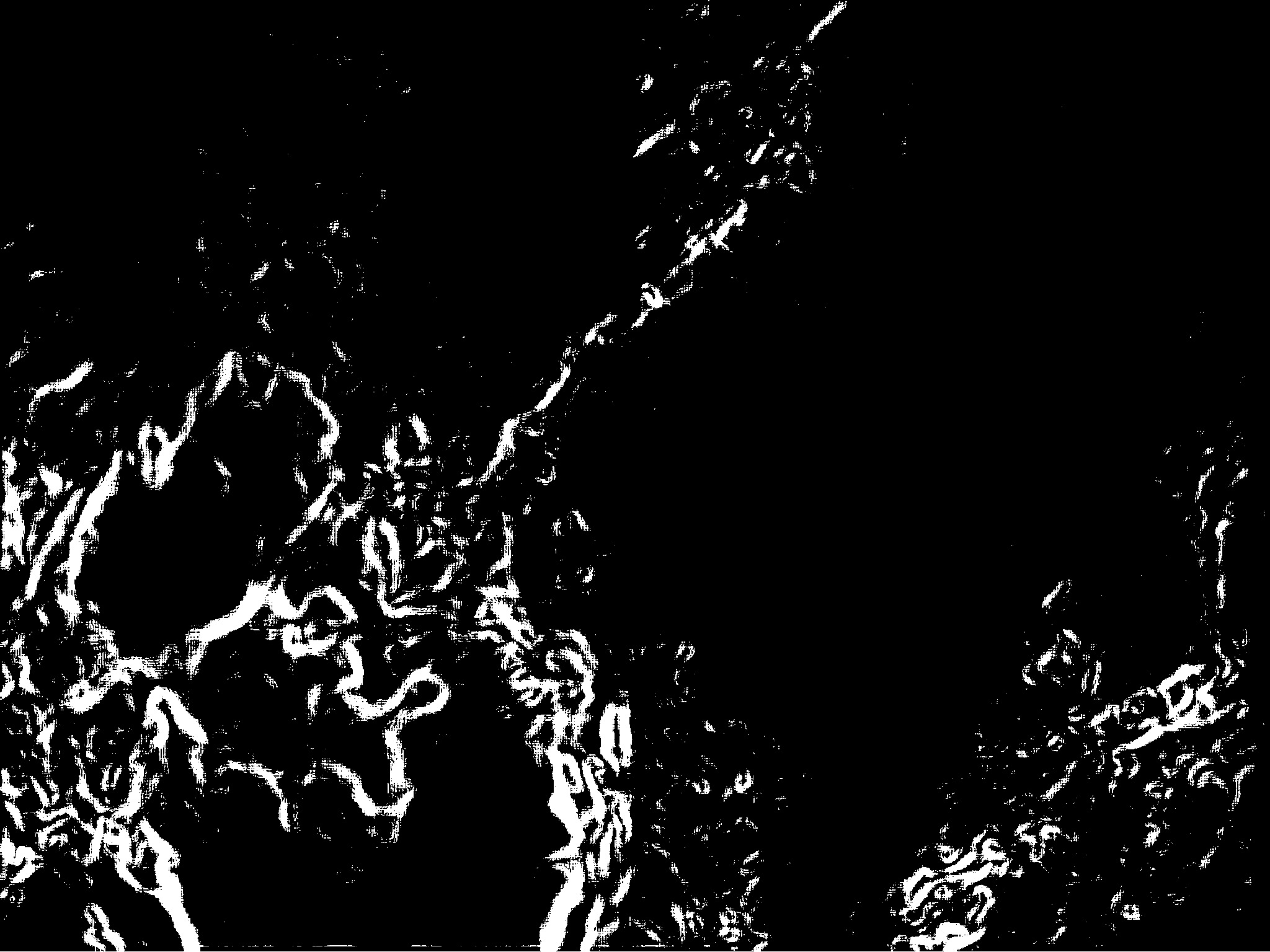}
        \end{subfigure}
        \caption{Border extraction from deblurred image}
        \label{fig:borders_deblurred}
    \end{subfigure}
    \caption{Edge detection using a Sobel filter from both the raw image (\ref{fig:borders_raw}) and the deblurred version of it (\ref{fig:borders_deblurred})}.
    \label{fig:borders}
\end{figure}

The proposed deblurring approach significantly enhances image clarity and structural reconstruction. Initial models trained on Sentinel-2 imagery were able to improve the sharpness of IMAGIN-e data (see Fig. \ref{fig:initial_model}), enabling subsequent georeferencing and a more comprehensive characterization of noise type, effective resolution, and spectral sensitivity. In addition, the application of a Sobel edge detection filter confirmed that, despite some undetected boundaries, the edges of critical objects and terrains were more clearly delineated (See Fig. \ref{fig:borders}). These improvements are paramount for subsequent object detection and segmentation tasks in onboard applications.

Quantitative evaluation demonstrates a substantial enhancement in image quality across multiple metrics (see Table \ref{tab:image_quality_metrics}). On Sentinel-2 images, SSIM improved by 72.47\% and PSNR increased by 25.00\%, calculated by comparing noisy synthetic images with reference images in the initial state and processed images with the same references in the final state. In contrast, for IMAGIN-e, image perceptual quality improved significantly, with NIQE showing a 60.66\% enhancement and BRISQUE improving by 48.38\%. Since these metrics evaluate image quality without requiring clean reference images, they are particularly valuable for real-world applications where reference-free assessment is necessary, as is the case for IMAGIN-e. 

From a computational standpoint, the deblurring process operates within the edge computing constraints outlined in Table \ref{tab:conditions}. Under these conditions, the model successfully processes a 2048x1536 pixel image in approximately 5 minutes, demonstrating its ability to handle high-resolution inputs despite resource limitations. Peak memory consumption reaches 600 MB, exceeding the available RAM and requiring the use of virtual memory. While this contributes to an extended processing time, the results highlight the model’s adaptability in constrained environments and underscore the role of efficient memory management in optimizing performance.

\begin{figure}[t]
    \centering
    \begin{tikzpicture}
        \node {\includegraphics[width=0.5\columnwidth]{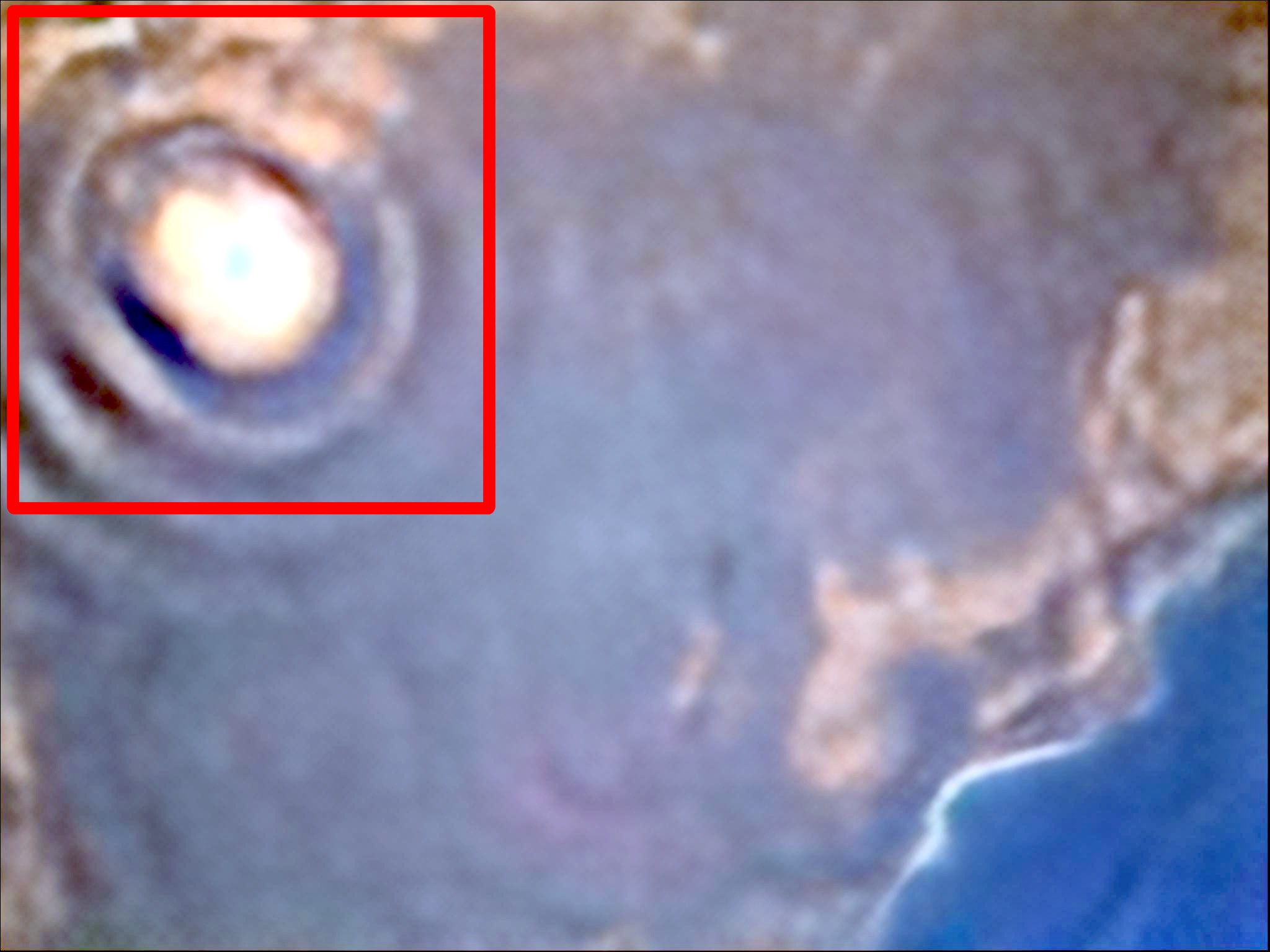}};
        \begin{scope}[xshift=2.5cm, yshift=0.5cm]
            \node {\includegraphics[width=2.7cm]{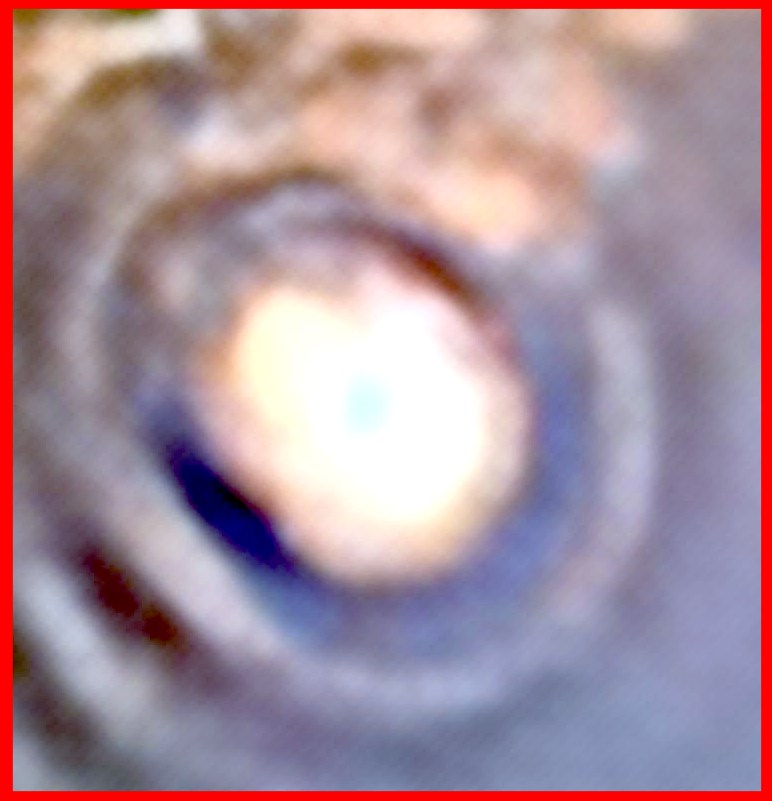}};
        \end{scope}
    \end{tikzpicture}
    \caption{Ringing effect on the images}
    \label{fig:ringing}
\end{figure}

Occasional ringing artifacts were observed, probably due to scaling operations during patch processing (see Fig.~\ref{fig:ringing}). Moreover, the effective Ground Sample Distance (GSD) varied between 37.4 m and 41 m, reflecting the dynamic imaging conditions of the ISS and underscoring the need for adaptive processing workflows.

\begin{table}[t]
    \caption{Image quality metrics for Sentinel-2 synthetic validation images and IMAGIN-e real ones}
    \label{tab:image_quality_metrics}
    \centering
    \begin{tabular}{|l|l|c|c|c|}
        \hline \hline
        \textbf{Dataset} & \textbf{Metric} & \textbf{Original} & \textbf{Deblurred} & \textbf{$\Delta$\%} \\
        \hline
        \multirow{2}{*}{\makecell{Sentinel-2 \\ (Synthetic)} } & SSIM & 0.4442 & 0.7662 & \textbf{+72.47\%}\\
        & PSNR & 24.0127 dB & 30.0159 dB & \textbf{+25.00\%}\\
        \hline
        \multirow{2}{*}{\makecell{IMAGIN-e \\ (Real)}} & NIQE & 21.9257 & 8.6263 & \textbf{+60.66\%}\\
        & BRISQUE & 110.8351 & 57.2149 & \textbf{+48.38\%}\\
        \hline \hline
    \end{tabular}
\end{table}


\section{Conclusions and Future Work}
\label{sec:conclusions}

Despite the inherent complexity of blind deblurring, our results demonstrate that incorporating Sentinel-2 imagery enables an effective iterative processing approach. This strategy allowed us to refine the image synthesis techniques and achieve acceptable outcomes—even without access to a sharp reference image. The final model is fast and efficient enough to be executed onboard during the post-processing phase, ensuring compatibility with the IMAGIN-e mission and maximizing the use of the instrument, which might otherwise be underutilized.

Moreover, the restored images prove valuable for specific applications, such as water body segmentation and coarse contour detection for map generation. However, it is important to note that while these results are promising for certain contexts, the current resolution is insufficient for detecting small objects or for the fine segmentation of closely related classes. This limitation reflects the trade-off between processing speed and image quality inherent in edge computing scenarios.

Further research could focus on leveraging enhanced onboard computational resources to deploy more powerful models that process image patches at their original resolution. By eliminating the need for downscaling and subsequent upscaling, this approach would likely yield images with increased realism and detail. Such improvements could enhance the deblurring performance while expanding the applicability of the processed imagery, especially in tasks requiring the detection of small objects or fine-grained segmentation.


\bibliographystyle{unsrtnat}
\bibliography{chapters/bibliography}

\small

\end{document}